\documentclass[11pt,a4paper]{article}
\usepackage{times,latexsym}
\usepackage[hyphens]{url}

\usepackage[T1]{fontenc}
\usepackage{CJKutf8}
\usepackage{adjustbox}
\usepackage{booktabs}
\usepackage{array}

\usepackage{authblk}
\usepackage{graphicx}
\usepackage{caption}
\usepackage{makecell}
\usepackage{subcaption}
\usepackage{microtype}

\usepackage{tikz}
\usepackage{xcolor}
\usepackage{longtable}
\usepackage{booktabs}
\usepackage{caption}
\usepackage{float}
\usepackage{placeins}
\usepackage{adjustbox}

\usepackage{tcolorbox}
\usepackage{enumitem}
\usepackage{lipsum}
\usepackage{geometry}
\usepackage{wrapfig}

\usepackage[utf8]{inputenc}
\usepackage{CJKutf8}

\usepackage[acceptedWithA]{tacl2021v1}

\usepackage[]{tacl2021v1}

\usepackage{xspace,mfirstuc,tabulary}

\usepackage{graphicx} 
\usepackage{xspace}
\newcounter{notecounter}
\newcommand{\enotesoff}{\long\gdef\enote##1##2{}}
\newcommand{\enoteson}{\long\gdef\enote##1##2{{
\stepcounter{notecounter}=
{\large\bf
\hspace{1cm}\arabic{notecounter} $<<<$ ##1: ##2
$>>>$\hspace{1cm}}}}}

\enoteson
\enotesoff

\title{
\approachnameshort: High-Quality Instruction Tuning Datasets for Low-Resource Languages via Reverse Instructions
 }

\def\approachname{Multilingual Reverse Instructions\xspace}
\def\approachnameshort{MURI\xspace}
\def\datasetname{\textsc{MURI-IT}\xspace}
\def\modelname{\textsc{MURI-101}\xspace}
\def\monomodelname{\textsc{MURI\textsubscript{1}}\xspace}
\def\datasetsize{2 million\xspace}
\def\languagecount{200\xspace}
\def\humanevallangnumber{13\xspace}

\def\culturaxdoccountbeforefilter{1,076,575\xspace} 
\def\culturaxdoccountafterfilter{686,723\xspace} 
\def\culturaxlangcount{123\xspace} 

\def\wikipediadoccountbeforefilter{1,554,207\xspace}
\def\wikipediadoccountafterfilter{1,031,726\xspace}
\def\wikipedialangcount{187}

\def\wikihowdoccount{54,578\xspace}
\def\wikihowlangcount{18\xspace}

\def\oasstdoccount{9,486\xspace}
\def\oasstlangcount{10}

\def\supnatdoccount{161,986\xspace}
\def\supnatlangcount{55}

\def\flandoccount{100,000\xspace}
\def\flanlangcount{1}

\def\xpdoccount{184,000\xspace}
\def\xplangcount{44}

\def\mmmlunumlangs{31\xspace}

\renewcommand*{\Affilfont}{\normalsize\normalfont}

\makeatletter
\renewcommand\AB@affilsepx{\hspace{1em} \protect\Affilfont}
\makeatother

\author[1,2,4]{Abdullatif Köksal}
\author[1]{Marion Thaler}
\author[1,2]{Ayyoob Imani}
\author[3]{Ahmet Üstün}
\author[4]{\authorcr Anna Korhonen}
\author[1,2]{Hinrich Schütze}

\affil[1]{CIS, LMU Munich}
\affil[2]{Munich Center for Machine Learning\protect\\\hspace{-0.5cm}}
\affil[3]{Cohere for AI}
\affil[4]{Language Technology Lab, University of Cambridge\protect\\\texttt{akoksal@cis.lmu.de}}

\date{}

\begin{document}
\maketitle
\begin{abstract}
Instruction tuning enhances large language models (LLMs) by
aligning them with human preferences across diverse
tasks. Traditional approaches to create instruction tuning
datasets 
face serious challenges for low-resource languages due 
to their dependence on data annotation. 
This work introduces a novel method,
\approachname{} (\approachnameshort{}), which generates
high-quality instruction tuning datasets for low-resource
languages without requiring human annotators or pre-existing
multilingual models. Utilizing reverse instructions and a
translation pipeline, \approachnameshort{} produces
instruction-output pairs from existing human-written texts
in low-resource languages. This method ensures
cultural relevance and diversity by sourcing texts from
different native domains and applying filters to eliminate
inappropriate content. 
Our dataset, \datasetname, includes more than 2 million
instruction-output pairs across \languagecount{}
languages. Evaluation by native speakers and fine-tuning
experiments with mT5 models demonstrate the approach's
effectiveness for both NLU and open-ended generation.
We publicly release datasets and models at
\href{https://github.com/akoksal/muri}{https://github.com/akoksal/muri}.
\end{abstract}

\section{Introduction}

Instruction tuning refines large language models (LLMs) based on user intentions, enhancing their ability to generalize across tasks and align with human preferences \citep{ouyang2022training, sanh2022multitask,muennighoff-etal-2023-crosslingual,wang-etal-2022-super}. While pre-training data can be automatically collected from the web, preparing instruction tuning data is challenging as it requires aligned instruction-output pairs. Three main approaches have been applied to create instruction tuning datasets: human annotation \citep{ouyang2022training, kopf2023openassistant, conover2023free}, templatized NLP tasks \citep{sanh2022multitask, wang-etal-2022-super, longpre2023flan}, and synthetic data generation via LLMs \citep{wang2023selfinstruct, honovich2022unnatural}.

For low-resource languages, these approaches face serious limitations. Human annotation is costly, and finding native speakers for low-resource languages is challenging. Templatizing NLP tasks restricts datasets to specific structures and domains, limiting their general applicability, and there is insufficient NLP task-annotated data for low-resource languages \citep{imanigooghari-etal-2023-glot500}. Synthetic data generation is constrained by the languages supported by existing models and suffers from validity \citep{wang2023selfinstruct} and creativity \citep{honovich2022unnatural} issues. Moreover, outputs of both template-based and synthetic data generation methods heavily rely on translation pipelines and are particularly prone to artifacts known as ``translationese'' \cite{gellerstam1986translationese}. These artifacts include simplified vocabulary and grammar, unidiomatic word order and expressions, and often neglect linguistic and cultural contexts. Such occurrences of translationese have been shown to negatively impact model training \cite{yu2022translate} by distorting examples and further distancing them from their linguistic and cultural context.

Consequently, no existing model, open source or proprietary,
supports
low-resource languages with the quality necessary for
high-quality instruction tuning dataset generation.
This has resulted in a disparity, with English
dominating 73\% of the most popular datasets \citep{longpre2023data}, leaving low-resource language communities underserved.

In this work, we introduce \approachname
(\approachnameshort), a novel approach to generate
instruction tuning datasets for low-resource languages
without requiring annotators, task-annotated data, or
pre-trained multilingual models.
\approachnameshort employs the reverse instructions method proposed
by \citet{koksal2024longform} and combines it with machine translation
to develop language-specific instructions,
$i_{\tau}$, for a text $d_{\tau}$. This involves translating $d_{\tau}$ to
English, generating an English instruction $i_{\epsilon}$ using reverse
instructions, and translating $i_{\epsilon}$ back to the
original language, so it can serve as instruction for 
 $d_{\tau}$. Notably, unlike fully translation-based methods, our approach requires only translating the instructions which enables using authentic outputs in the target language. The instruction-output pair ($i_{\tau}$, $d_{\tau}$) can then be used 
to fine tune a language model to follow instructions.
This approach is cost-effective 
and applicable to any language with available textual data.

By applying \approachnameshort to texts from diverse sources, 
we have created \datasetname, a dataset containing more than \datasetsize{} 
instruction-output pairs across \languagecount{} languages. To our 
knowledge, this dataset offers the broadest language coverage for 
multilingual instruction tuning. Our sources include Wikipedia, 
WikiHow, and various web-crawled pages, providing a rich variety in 
style, domain, and length. The output documents, sourced directly 
from data in their original languages, retain cultural and linguistic 
nuances. Additionally, we use quality filters to ensure the dataset's 
high standards.

To evaluate \datasetname, native speakers across \humanevallangnumber 
languages assess the dataset on five aspects to gauge quality. 
We also fine-tune several mT5 family models using \datasetname to execute 
instruction-based tasks, assessing their performance in natural language 
understanding and generation. For instance, \modelname, an mT5-XXL model 
instruction-tuned
with \datasetname,
outperforms prior models
like mT0 \citep{muennighoff-etal-2023-crosslingual} by
over 14\% in multilingual MMLU. In open-ended
generation tasks, it delivers much better
outputs than mT0, with win rates of 59\% vs.\ 28\%. Additionally, \datasetname
enhances performance when used alongside existing datasets like Aya.
We make the fine-tuned models and \datasetname publicly available.

To summarize, our contributions are:\\
    (i) We introduce \approachname (\approachnameshort),
      a cost-effective method for creating multilingual instruction
      tuning datasets applicable to hundreds of languages.
       \\
    (ii) We create and publish \datasetname, an instruction tuning 
    dataset for \languagecount{} languages using \approachnameshort. 
    This dataset consists of 2,228,499 instruction-output pairs, with 64\% of the data 
    from low-resource languages.\\
    (iii) We evaluate and analyze the dataset with native
      speakers in \humanevallangnumber languages. We find that the data
      is highly idiomatic in many languages. \\
    (iv) We fine-tune and release \modelname, a massively
      multilingual instruction-following language model
      using \datasetname.

\section{Related Work}

\paragraph{Instruction Tuning Datasets}
Instruction tuning has emerged as a powerful approach to
enhancing the instruction-following capabilities of LLMs, as
demonstrated by numerous studies \citep{ouyang2022training,
  sanh2022multitask, muennighoff-etal-2023-crosslingual,
  wang-etal-2022-super}. The three primary strategies to create
instruction tuning datasets
are human curation, templatized tasks, and synthetic generation via LLMs.

Human-curated datasets, like Open Assistant
\citep{kopf2023openassistant} and Dolly
\citep{conover2023free}, involve extensive human annotation
but are difficult to scale and extend to more languages due to high cost.
Alternatively, datasets such as Public Pool of Prompts (P3)
\citep{sanh2022multitask}, SuperNatural Instructions (NIv2) \citep{wang-etal-2022-super},
and FLAN \citep{longpre2023flan} utilize NLP task reformulation to reduce cost
and enhance applicability but still struggle with general-purpose
 instruction following since their main focus is on NLU tasks.

To address these issues, synthetic datasets have been developed, 
such as Self-Instruct \citep{wang2023selfinstruct}, TeGit \citep{chen2023tegit}, 
Unnatural Instructions \citep{honovich2022unnatural}.
These datasets offer greater task diversity but are 
challenged by issues of validity and creativity.
The reverse instruction method \cite{koksal2024longform},
employing data augmentation via generative models and pretraining corpora,
further exemplifies
cost-effective dataset generation.
A similar method has been successfully applied in Bactrian-X \citep{li2023bactrianx},
demonstrating its effectiveness in multilingual settings by leveraging translation
for synthetic data.

\begin{figure*}[t]
	\centering
	\includegraphics[width=0.91\textwidth]{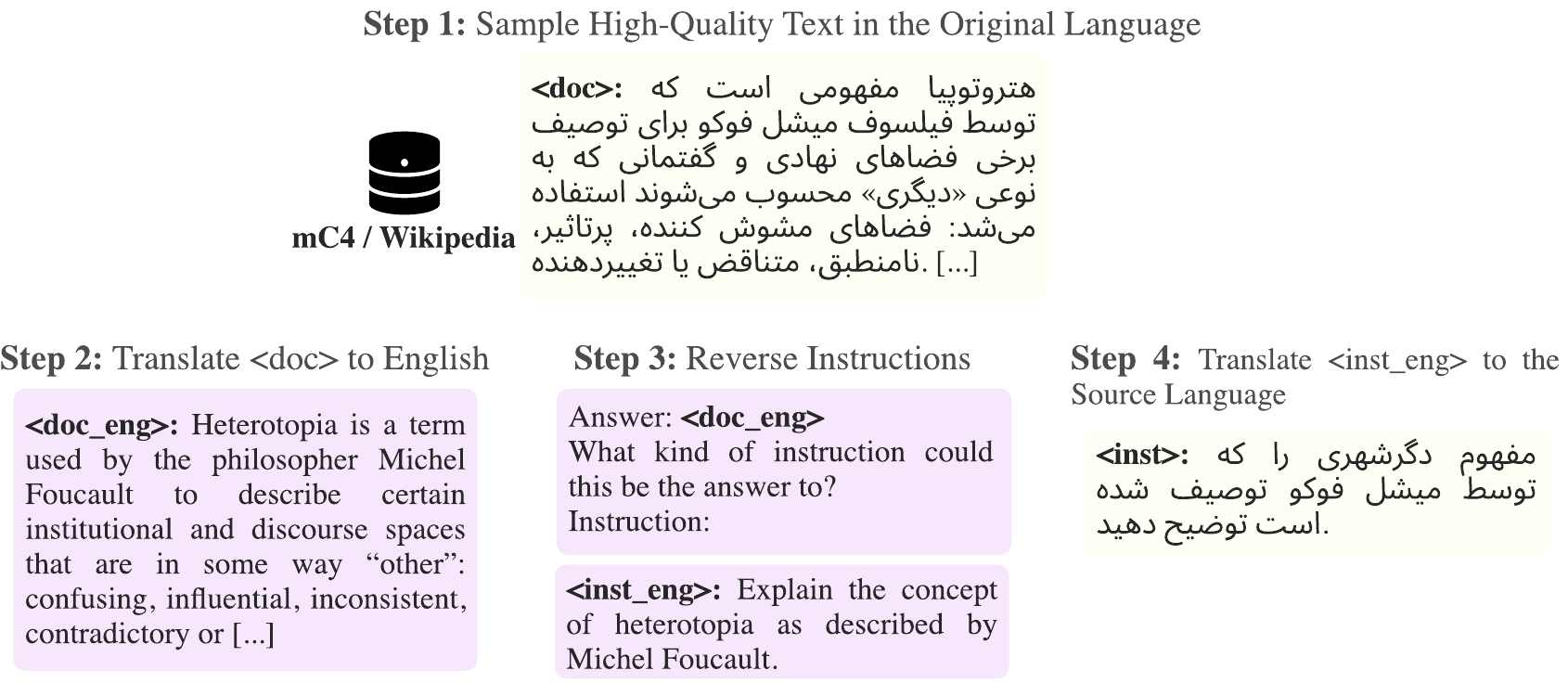} 
	\caption{%
		\textbf{\approachname}
		(\approachnameshort{}). Step 1: \approachnameshort{}
		selects a high-quality human-written example (<doc>) from
		multilingual corpora. Step 2: Translation  into
		the English document <doc\_eng>. Step 3 applies the reverse instructions
		method to <doc\_eng> (i.e., prompting the LLM to  generate a matching
		instruction <inst\_eng>). Step 4: <inst\_eng>
		is translated back into the source language
		(<inst>), resulting in a (<inst>, <doc>) pair where
		the <doc> output is human-written.}
	\label{fig:muri_workflow}
\end{figure*}

Multilingual instruction tuning has shown substantial benefits,
especially for low-resource languages
\citep{muennighoff-etal-2023-crosslingual}. It
not only maintains performance in English but also
enhances capabilities in non-English languages with
the help of a large scale of English examples
\citep{shaham2024multilingual, llama3modelcard}.
Despite these
advancements, large-scale multilingual datasets often remain
limited. Efforts to overcome this include pre-training on
diverse multilingual data \citep{chung2022scaling,
  chowdhery2022palm} and creating dedicated multilingual
instruction fine-tuning sets
\citep{muennighoff-etal-2023-crosslingual}, and manual
annotation by native speakers
\citep{singh2024aya}. Extensions to existing datasets often
utilize automatic translation \citep{li2023bactrianx,
  winata-etal-2023-nusax,holmstrom-doostmohammadi-2023-making}
and template-based generation \citep{gupta2023targen}. These
methods strive to balance diversity against resource
constraints and quality issues inherent in automated
translation processes, as seen in the extensive Aya
collection \citep{singh2024aya}.
In summary, while instruction tuning has greatly advanced the capabilities of
LLMs in following complex instructions, challenges remain in dataset diversity,
validity, and the integration of multilingual content.

\paragraph{Multilingual LLMs}
In the recent surge of LLMs, English-centric models like the 
closed-source GPT model family \citep{radford2019language, ouyang2022training} 
and open-sourced ones like LLaMA and Pythia 
\citep{touvron2023llama, biderman2023pythia}, have gained prominence.  
Multilingual models, unlike monolingual ones, offer the advantage of facilitating 
cross-lingual tasks such as translation \citep{jiao2023parrot,xu2024paradigm} 
and addressing low-resource languages 
\citep{artetxe2019massively, wu-dredze-2020-languages}. 
mBERT pioneered the multilingual area by demonstrating that training on 
multilingual data allows different languages to be represented in a unified 
semantic space
\cite{devlin-etal-2019-bert}. Building on this foundation, subsequent models
such as XLM-R, GLoT500, and AfriBERTa extended the capabilities of
transformer-based models to  hundreds of languages
\cite{conneau-etal-2020-unsupervised, imanigooghari-etal-2023-glot500, ogueji-etal-2021-small}.

However, the progression of multilingual encoder-decoder and 
decoder-only models has been more restrained.  
Models like LLaMA and Mistral, for instance, feature datasets 
predominantly composed of English, with limited data from a select 
group of high-resource languages 
\cite{touvron2023llama, jiang2023mistral}. 
In contrast, models like XGLM,
 BLOOM, and MGPT have been developed from scratch to support extensive 
 language diversity 
 \cite{workshop2023bloom, lin2022fewshot, oleh-2022-memgpt}. 
 Meanwhile, mT5 is trained on 101 languages,
 an important step
in
 encoder-decoder model training \cite{xue-etal-2021-mt5}.

\section{The \datasetname Dataset}

We introduce \datasetname{}, which includes
2,228,499 instruction-output pairs in 200 languages. The
dataset is
primarily
constructed by applying \approachname{}
(\approachnameshort{}) to the CulturaX
\cite{nguyen2023culturax} and Wikipedia corpora. The core
idea is summarized in Figure \ref{fig:muri_workflow}. Our
goal is to utilize existing high-quality human-written
multilingual corpora to generate a diverse
instruction-following dataset.
For a randomly selected text, we aim to generate an
instruction for which the high-quality corpus text
would serve as a good response.
This approach ensures that a model trained with this dataset
will not be conditioned on outputs with translation artifacts or culturally irrelevant topics.

\datasetname{} also incorporates two additional subsets:
\wikihowdoccount{} instances collected from the WikiHow
website
in \wikihowlangcount{}
languages to augment the dataset, and 455,472 existing NLP task examples
in 74 languages to enrich its diversity.
This section details the steps used to produce \datasetname{}.

\subsection{Multilingual Reverse Instructions (MURI)}

\textbf{Step 1. Data Selection:}
We randomly sample documents from two multilingual corpora:
CulturaX (\culturaxdoccountbeforefilter{} documents) and
Wikipedia (\wikipediadoccountbeforefilter{} documents).
CulturaX encompasses 167 languages, merging the OSCAR
\cite{ortiz-suarez-etal-2020-monolingual} and mC4
\cite{xue-etal-2021-mt5} corpora with additional cleaning,
deduplication, language identification, and diversification
procedures \cite{nguyen2023culturax}. Wikipedia spans over 350
languages with high-quality documents.

\noindent{\textbf{Instruction Generation}}
After selecting high-quality outputs, the next step is
generating suitable instructions in the source
language. Since recent LLMs support a limited number of
languages, we utilize machine translation models and English
LLMs.
Let $(i_{\tau}^k,
    d_{\tau}^k)$ be  an instruction-output pair in
    a target low-resource language $\tau$.
Given a corpus of human-written documents $D_{\tau} = {d_{\tau}^1, d_{\tau}^2, \ldots, d_{\tau}^n}$,
we aim to create
an instruction tuning dataset $D_{I\tau} = {(i_{\tau}^1,
  d_{\tau}^1), (i_{\tau}^2, d_{\tau}^2), \ldots,
  (i_{\tau}^n, d_{\tau}^n)}$.

\textbf{Step 2. Document Translation:}
First, each document $d_{\tau}^k$ is translated to English using a machine translation model, resulting in $d_{\epsilon}^k$. We use MADLAD-400-3B-MT \cite{kudugunta2023madlad400}, with top\_p=$1$ sampling for translation.

\textbf{Step3. Reverse Instructions:}
Next, we employ an English LLM for instruction generation. We 
modify the reverse instructions prompt in \citep{koksal2024longform}
to generate an instruction $i_{\epsilon}^k$ for $d_{\epsilon}^k$ in a 
few-shot manner, as illustrated in Table \ref{tab:few_shot_prompt} in Appendix.
We use Mixtral-8x7B \cite{jiang2024mixtral} with greedy decoding for instruction generation.

\textbf{Step 4. Translating Instruction to the Source Language and Ensuring Language Consistency:}
Finally, the generated instruction $i_{\epsilon}^k$ is translated back to its source language using MADLAD-400-3B-MT, denoted as $i_{\tau}^k$. To verify language consistency, we utilize GlotLID \cite{kargaran-etal-2023-glotlid} and discard mismatched translations.

\textbf{Step 5. Content screening:}
We observe that some examples contain violent or noisy content due to the nature of the corpora. We utilize the RoBERTa hate-speech model \cite{vidgen2021lftw} to screen the generated instruction-output $(i_{\epsilon}^k, d_{\epsilon}^k)$ pairs in English and eliminate unsuitable examples. To ensure dataset integrity and eliminate redundancy, we employ the MinHashLSHForest method for deduplication.

Manual screening revealed unsuitable instructions lacking
necessary context, such as instructions asking for
summarization of non-existent prior documents or requesting
translations. We excluded these instruction-output pairs
from our dataset by filtering out instructions including the
words \textit{summarize} or
\textit{translate}. Additionally, we observed that
web-sourced documents often include extraneous content like
footers, headers, or advertisements. We leave the
elimination of such
extraneous content for future work.

\subsection{WikiHow Data}
We collected articles from the multilingual WikiHow website
using PyWikiHow
\cite{pywikihow}
in \wikihowlangcount{} languages (Arabic, Chinese, Czech, Dutch, English, French, German, Hindi, Italian, Japanese, Korean, Malay, Portuguese, Russian, Spanish, Thai, Turkish, and Vietnamese), based on the URLs provided by Wikilingua \cite{ladhak-etal-2020-wikilingua}. Each WikiHow page is comprised of the following sections:
(i) A \textit{title} that starts with ``How to'', (ii) an \textit{abstract} answer to the question, (iii) a number of \textit{steps}, each comprised of a \textit{step-title} and a \textit{step-text} paragraph.
We use the \textit{title} of each WikiHow page as the instruction. To introduce variation in the style of the answers, we render the answers to the questions as follows:
In $50\%$ of cases, we include the \textit{abstract} in the answer and in the other $50\%$ we don't. Regardless of whether the \textit{abstract} is included or not, in $50\%$ of the cases we only include the \textit{step-titles}, and in the other
$50\%$ we include both the \textit{step-titles} and \textit{step-texts}.

\subsection{NLP Tasks}
To further improve diversity of tasks in \datasetname{}, we incorporated several existing multilingual instruction following datasets based on NLP tasks, expanding language coverage and task diversity. These additions, totaling 455,472 samples across 74 languages, complement our primary data sources:

\textbf{SuperNatural Instructions}: We sampled 200 tasks from SuperNatural Instructions \citep{wang-etal-2022-super} per translation task type and 500 samples from the remaining set, resulting in \supnatdoccount{} samples across \supnatlangcount{} languages.

 \textbf{xP3} comprises 46 languages across 16 NLP tasks, such as various types of QA and Topic Classification. 
 We adapted 184,000 samples of xP3 \cite{muennighoff2022crosslingual} to our format.

\textbf{OASST1}: From this crowd-sourced chat-style dataset \cite{kopf2024openassistant} spanning 35 languages, we selected and paired message and output pairs up to the second deepest level, yielding 9,486 examples in 10 languages.

 \textbf{FLAN v2}: We incorporated 100,000 samples from FLAN v2 \citep{longpre2023flan}, including 50,000 from its main collection and 50,000 from its Chain-of-Thought subset, all in English following Tulu \citep{ivison2023camels}.

These additional datasets were chosen to increase the linguistic and 
task diversity of \datasetname{}, ensuring a more comprehensive 
and versatile instruction-tuning dataset similar to prior work like Aya 
\citep{singh2024aya}.
We have summarized the statistics of our dataset in 
Table \ref{tab:data_stats}.

\begin{table}[t]
    \centering

    \resizebox{\linewidth}{!}{
    \begin{tabular}{lrr}
        \toprule
        \textbf{Source} & \textbf{\# Languages} & \textbf{\# Examples} \\
        \midrule
        \textbf{Multilingual Reverse Instructions} & 194 & 1,718,449 \\
        \midrule
        \quad Wikipedia & \wikipedialangcount{} & \wikipediadoccountafterfilter{} \\
        \quad CulturaX & \culturaxlangcount{} & \culturaxdoccountafterfilter{} \\
        \midrule
        \midrule
        \textbf{WikiHow} & \wikihowlangcount{} & \wikihowdoccount{} \\
        \midrule
        \midrule
        \textbf{NLP Tasks} & 74 & 455,472 \\
        \midrule
        \quad SupNatInst-v2 & \supnatlangcount{} & \supnatdoccount{} \\
        \quad xP3 & \xplangcount{} & \xpdoccount{} \\
        \quad OpenAssistant & \oasstlangcount{} & \oasstdoccount{} \\
        \quad FLAN v2.0& \flanlangcount{} & \flandoccount{} \\
        \midrule
        \midrule

        \textbf{Total} & \textbf{200} & \textbf{2,228,499} \\
        \bottomrule
    \end{tabular}
    }
    \caption{Composition of \datasetname{}  by
      source, including the number of languages and
      examples. We split the dataset 90/5/5 into
      training, validation, and test sets,
      ensuring similar ratios for sources and
      languages.}
    \label{tab:data_stats}
\end{table}

\section{Dataset Analysis}
This section presents an in-depth analysis of \datasetname,
focusing on two key aspects: linguistic diversity and data quality assessment.
Section \ref{Languages and linguistic diversity} examines the range of languages
represented in \datasetname, highlighting its coverage of low-resource languages,
diverse scripts, word orders, and case-marking systems.

Section \ref{sec:quality_assessment} evaluates the dataset's effectiveness in
maintaining linguistic and cultural accuracy. We detail the findings from a quality
evaluation involving native speakers, assessing alignment, correctness, and overall
informational sufficiency.

\begin{figure*}[t]
    \centering
    \begin{minipage}[b]{0.49\textwidth}
        \centering
        \includegraphics[width=\columnwidth]{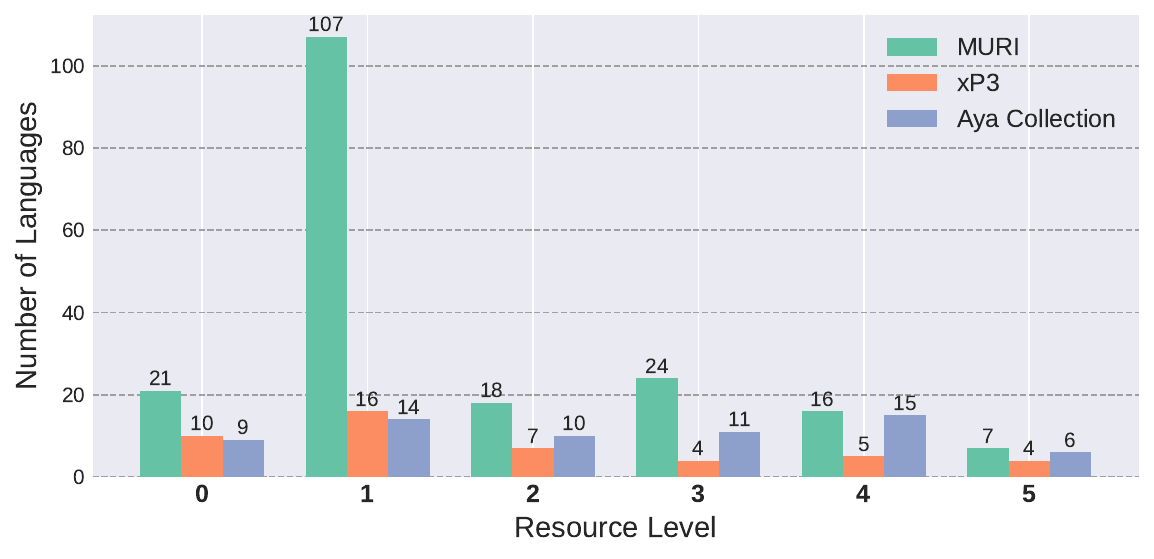}
        \subcaption{Resource levels of languages in \datasetname, xP3, and Aya datasets (per \citet{joshi2021state}).}        
        \label{fig:languages_per_resource_level}
    \end{minipage}%
    \hfill
    \begin{minipage}[b]{0.49\textwidth}
        \centering
        \includegraphics[width=\columnwidth]{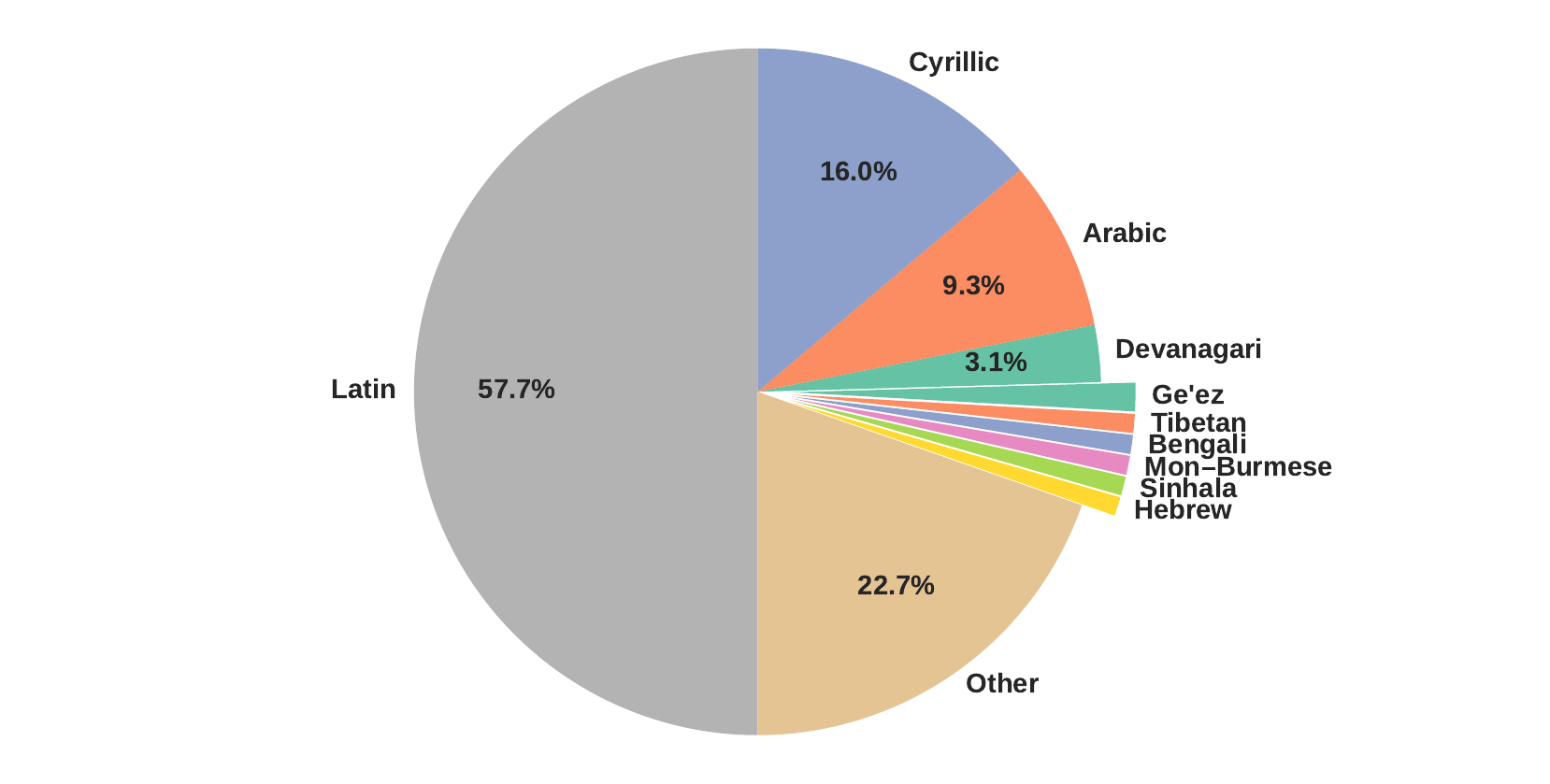}
        \subcaption{Percentage of languages in \datasetname by script.}
        
        \label{fig:languages_per_script}
    \end{minipage}

    \vspace{0.01cm} %

    \begin{minipage}[b]{0.49\textwidth}
        \centering
        \includegraphics[width=\columnwidth]{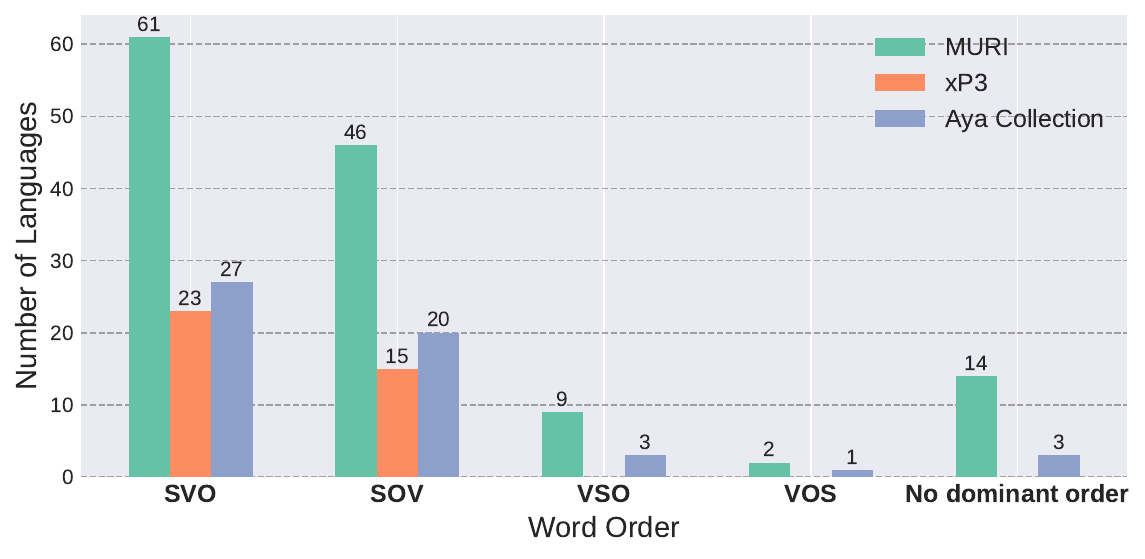}
        \subcaption{Word order in \datasetname, xP3, and Aya datasets (per \citet{wals-81}).}
        \label{fig:word_orders}

    \end{minipage}%
    \hfill
    \begin{minipage}[b]{0.49\textwidth}
        \centering
        \includegraphics[width=\columnwidth]{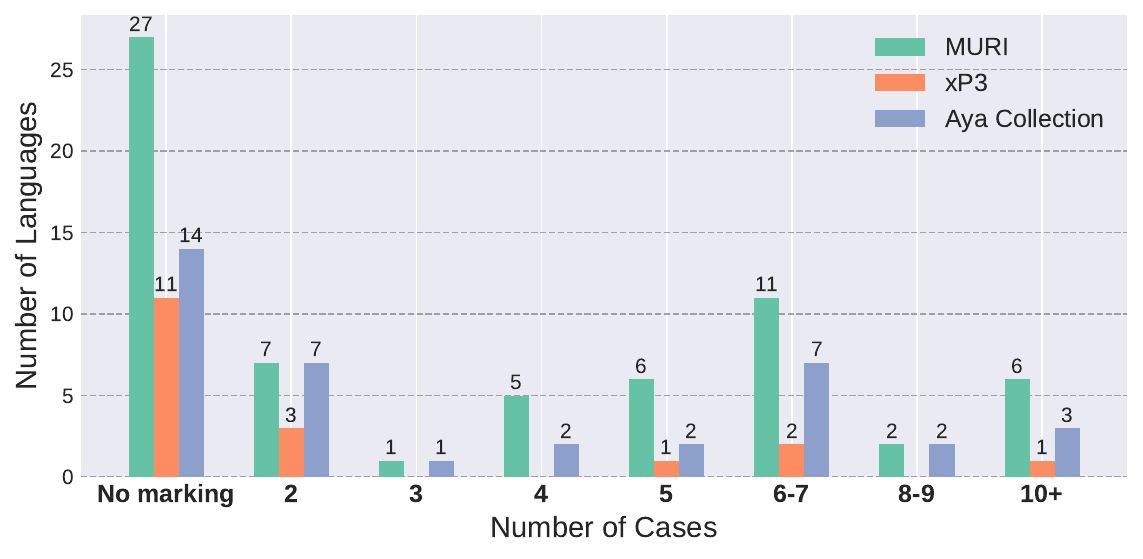}
        \subcaption{Case marking in \datasetname, xP3, and Aya datasets (per \citet{wals-49}).}
        
        \label{fig:cases}
    \end{minipage}

    \vspace{-0.02cm} %

    \caption{Linguistic diversity of \datasetname compared to Aya \cite{singh2024aya} and xP3 \cite{muennighoff-etal-2023-crosslingual} datasets, highlighting differences in (a) resource level \cite{joshi2021state}, (b) script, (c) word order, and (d) case marking \cite{wals}. The classifications by \citet{joshi2021state} and WALS are extensive but not exhaustive, thus not covering all languages in the datasets.}

    \label{fig:combined_figure}
\end{figure*}
\subsection{Languages and Linguistic Diversity}
\label{Languages and linguistic diversity}

\approachnameshort{} aims to provide a methodology inclusive of low-resource languages through a culturally respectful approach, utilizing materials in their native languages and avoiding outputs in translationese \cite{bizzoni-etal-2020-human, vanmassenhove-etal-2021-machine}. Given that the majority of languages used in NLP systems share typological similarities and geographical origins \cite{joshi2021state}, this often leads to an uneven distribution of resources and tools available to the global community. \datasetname therefore focuses particularly on languages with limited resources and diverse features.

\citet{joshi2021state} outlined a taxonomy categorizing languages based on their resource levels, 
ranging from 0 (\textit{left-behinds}) such as Balinese with severely limited resources, 
to 5 (\textit{winners}) like English or French. Our dataset encompasses a large number of 
low-resource languages, as shown in Figure \ref{fig:combined_figure}.\subref{fig:languages_per_resource_level}, 
with over 700,000 examples falling into category 1. 
Despite this, access to outputs for these low-resource languages remains limited, 
with 33 languages containing fewer than 1,000 examples each. 
Nonetheless, \datasetname proves to be one of the most diverse instruction-tuning datasets to date.

\citet{nezhad2024exploring} emphasize script  and word order as important factors in analyzing
linguistic diversity. While the majority of languages in \datasetname employ
Latin script or a combination of Latin, Arabic, and Cyrillic scripts
(Figure \ref{fig:combined_figure}.\subref{fig:languages_per_script}),
a notable portion (more than one-fifth, categorized as ``Other'') features
low-resource scripts such as Lao or Georgian.
As the output texts have not been translated,
idiomatic use of these scripts is assured, ensuring correct orthography.

To further investigate linguistic diversity, we examined
word order and case marking.
Focusing on the order of subject, verb, and object,
Figure
\ref{fig:combined_figure}.\subref{fig:word_orders} shows
that
while European SVO languages predominate \cite{wals-81} and
there are no rare
OVS and OSV languages,
all frequent patterns
are represented.
This
showcases the
``structural'' diversity of our dataset.

Case-marking patterns align with geographical distribution; e.g.,
mid-size to large
inventories are prevalent in South Asia, Eastern Europe, and east-central
Africa \cite{wals-49}. Figure
\ref{fig:combined_figure}.\subref{fig:cases} illustrates
that our dataset encompasses a diverse range of case
systems, including complex systems with up to ten
cases. This
indicates that our dataset has good coverage of both
``analytic'' and ``synthetic'' languages.
Overall, case marking and word order exemplify
the broader coverage of \datasetname of less common
languages compared to previous datasets, contributing to a
more comprehensive representation of linguistic diversity in
NLP resources.

\begin{table*}[t!]
	\centering
	\scriptsize
	\begin{adjustbox}{max width=0.87\textwidth}
		\begin{tabular}{lccccc}
			\toprule
			\textbf{Language} & \textbf{Alignment} & \textbf{Prop. of Non-Instructions} & \textbf{Instruction Correctness} & \textbf{Output Correctness} & \textbf{Informational Sufficiency} \\
			\midrule
			English    & 4.70 & 0.00 & 4.97 & 4.70 & 4.27 \\
			German     & 4.73 & 0.07 & 4.57 & 4.77 & 4.03 \\
			Italian    & 4.16 & 0.07 & 4.16 & 4.26 & 4.23 \\
			Swedish    & 4.50 & 0.00 & 4.97 & 4.60 & 3.93 \\
			Dutch      & 4.37 & 0.10 & 4.90 & 4.33 & 4.20 \\
			French     & 4.53 & 0.13 & 4.80 & 4.53 & 4.43 \\
			Persian    & 4.00 & 0.03 & 4.53 & 4.60 & 4.00 \\
			Bavarian   & 3.47 & 0.53 & 3.90 & 4.63 & 4.30 \\
			Vietnamese & 4.47 & 0.07 & 4.50 & 4.50 & 3.33 \\
			Turkish    & 3.47 & 0.03 & 4.93 & 4.70 & 4.57 \\
			Chinese    & 3.37 & 0.10 & 5.00 & 4.23 & 4.37 \\
			Ukrainian  & 3.50 & 0.06 & 4.50 & 3.00 & 3.00 \\
			Korean     & 2.86 & 0.00 & 4.73 & 4.53 & 3.90 \\
			\midrule
			Avg. & 4.01 &	0.09 &4.65 & 4.41 & 4.04 \\
			
			\bottomrule
		\end{tabular}
	\end{adjustbox}
	\caption{Comparison of alignment, proportion of non-instructive phrasing, grammatical correctness, and informational sufficiency across \humanevallangnumber evaluated languages in the reverse instruction subset of \datasetname.}
	\label{tab:alignment}
\end{table*}

\begin{CJK*}{UTF8}{gbsn}
	\begin{table*}[t]
		\centering
		\resizebox{0.92\textwidth}{!}{
			\begin{tabular}{p{7.2cm}p{1.5cm}p{10cm}}
				\toprule
				\textbf{Issue} & \textbf{Lang.} & \textbf{Instruction \& Output} \\
				\midrule
				\textbf{No Orthographic Standard}: Standard German words in the
				instruction.
				Bavarian has no standard orthography, resulting in
				translation models mixing  different types of
				Bavarian and Standard German.
				& Bavarian & I: Beschreibn Sie die
				Geschichte von Pädaschduaf. \newline A: Pädaschduaf oda Pädasduaf (kuaz: P’dorf), offiziö Perchtoldsdorf, is a Moaktgmoa mit Einwohna (Stand ) im Bezirk Mödling in […] \\
				\midrule
				\textbf{Code-switching}: The instruction is in standard
				Mandarin, the output is in Shanghai dialect. & Chinese & I:
				犹太人是谁 \newline A: 犹太人（希伯来文: Yehudim）是一个历史悠久，凝聚力强的族群。[...] \\
				\midrule
				\textbf{Idiomaticity}: The expression
				\textit{Chronisch-Übersättigung} is not idiomatic
				(indicating shortcomings of the
				translation model).  & German &  I: Was ist chronische
				Übersäuerung des Körpers? Was bedeutet die
				Chronisch-Übersättigung des Organismus? \newline A: Die chronische übersäuerung des Körpers - Die chronische übersäuerung des körpers-Tagebücher Wenn der Körper  [...] \\
				\midrule
				\textbf{Lack of specific product information}: The output
				lacks specific information about  the tool \textit{TimeTool}
				that is required for a good output. & Dutch & I: Hoe kan ik
				mijn verjaardag vieren met TimeTools? \newline A: TimeTools heeft voor elk tijdzegel een uniek werkboek ontwikkeld. Veel mensen ervaren dit als een 'feest der herkenning'. […] \\
				\midrule
				\textbf{Superfluous Text Markup}: The output contains header
				details, inhibiting readability: \textit{mandy quotes billy
					and mandy |} & English & I: What is Clickfunnels? \newline
				A: mandy quotes billy and mandy | What is Clickfunnels? mandy quotes billy and mandy Prices: Is it truly worth it?\newline The Internet […] \\
				\bottomrule
			\end{tabular}
		}
		\caption{Examples of issues identified in the human evaluation of \datasetname, illustrating various challenges such as orthographic inconsistencies, code-switching, idiomatic inaccuracies, and superfluous text markup. These examples highlight areas where translation and formatting may impact the overall quality and clarity of the dataset.}
		\label{tab:issues}
	\end{table*}
\end{CJK*}

\subsection{Quality Assessment of \datasetname}
\label{sec:quality_assessment}

A distinctive feature of \datasetname{} is its preservation
of cultural and linguistic nuances, often lost in translated
datasets. To enhance our linguistic analysis, we conducted a
thorough evaluation of a random subset of the dataset,
involving native speakers proficient in
\humanevallangnumber{} languages. Each annotator examined 30
randomly selected instruction-output pairs from the reverse
instruction subset of \datasetname{} using five predefined
evaluation criteria. These criteria
assess the quality of both
instructions and outputs using -- except for Proper
Instruction Format -- a Likert scale.
(i)     \textbf{Alignment} (range 1-5): Measures the alignment between instruction and output.
   (ii)  \textbf{Instruction Correctness},
          (iii) \textbf{Output Correctness} (range 1-5): Assess
          lexical and grammatical accuracy of instruction
          and output.
   (iv)  \textbf{Informational Sufficiency} (range 1-5): Determines whether the instruction
     can be adequately answered without external context.
     (v) \textbf{Proper Instruction Format} (0: No, 1: Yes): Indicates whether the instruction is appropriately formatted for a language model.

Table
\ref{tab:alignment}
shows that  human assessment is generally good,
but with some
mixed results. High-resource
languages such as English, German, French and Italian
consistently perform well across all criteria. However, common 
issues include the presence of highly-specific and
ambiguous information and instructions that depend heavily on temporal context, which can reduce their clarity and usefulness in instruction tuning.
Across all 13 languages,
extraneous headers, footers, and metadata
are found in some outputs. Thus,
the noise contained in the underlying multilingual corpora affects
quality and coherence of \datasetname, as reflected in the slightly lower
average output correctness compared to input correctness. A
relatively minor problem is
less idiomatic and culturally
appropriate language use (see German example in Table
\ref{tab:issues}).

\begin{table*}[t]
	\centering
	\resizebox{\textwidth}{!}{
		\begin{tabular}{@{}lllllllllllllllll@{}}
			\toprule
			& \textbf{arb} & \textbf{ben} & \textbf{cat} & \textbf{dan} & \textbf{deu} & \textbf{eus} & \textbf{fra} & \textbf{guj} & \textbf{hin} & \textbf{hrv} & \textbf{hun} & \textbf{hye} & \textbf{ind} & \textbf{ita} & \textbf{kan} & \textbf{mal} \\
			\midrule
			\textsc{Okapi} & 27.7 & 26.8 & 30.5 & 31.8 & 31.7 & 27.9 & 30.7 & 27.4 & 26.5 & 30.0 & 30.1 & 27.5 & 27.5 & 30.4 & 26.8 & 25.8 \\
			\textsc{mT0} & 31.5 & 31.6 & 32.8 & 33.0 & 32.7 & 29.7 & 32.1 & 29.5 & 32.0 & 31.1 & 32.3 & 28.4 & 33.3 & 32.4 & 30.9 & 28.6 \\
			\textsc{mT0x} & 31.6 & 30.2 & 32.6 & 32.0 & 32.5 & 29.2 & 32.7 & 28.5 & 31.6 & 31.1 & 31.7 & 26.7 & 32.3 & 31.3 & 28.9 & 26.7 \\
			\textsc{Aya-101} & 38.2 & 35.8 & 39.6 & 39.7 & 39.7 & 36.0 & 39.7 & 33.6 & 38.7 & 37.5 & 38.8 & 30.0 & 40.0 & 39.0 & 34.5 & 30.4\\
			\modelname (ours) & 36.5 & 33.0 & 38.8 & 38.4 & 38.9 & 34.4 & 39.0 & 33.1 & 35.4 & 37.0 & 38.1 & 29.9 & 38.9 & 38.5 & 32.4 & 30.9 \\

			\midrule
			& \textbf{mar} & \textbf{nep} & \textbf{nld} & \textbf{por} & \textbf{ron} & \textbf{rus} & \textbf{slk} & \textbf{spa} & \textbf{srp} & \textbf{swe} & \textbf{tam} & \textbf{tel} & \textbf{ukr} & \textbf{vie} & \textbf{zho} & \textbf{Avg.} \\
			\midrule
			\textsc{Okapi} & 26.1 & 25.2 & 31.1 & 30.1 & 30.9 & 30.6 & 30.2 & 30.9 & 30.4 & 29.3 & 26.0 & 25.9 & 31.6 & 27.5 & 28.2 & 28.8 \\
			\textsc{mT0} & 31.6 & 32.4 & 32.0 & 32.1 & 32.4 & 32.8 & 32.3 & 32.1 & 30.9 & 31.6 & 29.4 & 29.0 & 31.5 & 30.9 & 32.5 & 31.5 \\
			\textsc{mT0x} & 29.7 & 30.1 & 32.1 & 32.0 & 31.8 & 31.7 & 31.4 & 32.2 & 31.4 & 32.8 & 27.7 & 27.9 & 32.3 & 31.1 & 31.6 & 30.8 \\
			\textsc{Aya-101} &  36.0 & 37.2 & 40.1 & 39.0 & 39.5 & 39.2 & 39.4 & 39.7 & 38.1 & 39.7 & 31.2 & 32.1 & 39.9 & 34.8 & 38.3 & 37.3 \\
			\modelname (ours) & 33.0 & 33.2 & 38.8 & 38.1 & 38.1 & 37.7 & 38.0 & 39.0 & 36.6 & 38.5 & 29.8 & 31.3 & 37.0 & 36.8 & 36.9 & 36.0 \\
			\bottomrule
		\end{tabular}
	}
	\caption{Multilingual MMLU performance of Okapi, mT0,
		mT0x, Aya-101 and \modelname{}  across 31
		languages. Scores are accuracy in a few-shot setup.
		Except for Okapi (25 shots), the number of shots is 5.}
	\label{tab:m_mmlu}
\end{table*}

For lower-performing languages, a major source of
error is the lack of standardization. For instance,
Bavarian -- spoken in Austria,
Bavaria and Alto Adige -- lacks a standard. This resulted in MADLAD
\cite{kudugunta2023madlad400} translations including Standard German
words (Table \ref{tab:issues}), degrading the quality of
generated instructions.
Similarly, Chinese instructions and outputs were sometimes
mismatched, e.g.,
inconsistent use of traditional vs.\ simplified Chinese and
of dialects vs.\
Standard Mandarin.

Overall, we observe a moderately high alignment between
instructions and outputs, averaging 4.01. Only 9\% of the
generated instructions deviate from the typical style of a
question or direct instruction.
Instruction-output pairs   are mostly grammatically and
lexically accurate, with higher-performing languages such as
English and German aligning particularly well. This directly
follows from
the superior performance of MADLAD
for these languages.

\section{Experimental Setup}
To evaluate the effectiveness of \datasetname{}, we instruction-tune mT5-XXL \citep{xue-etal-2021-mt5}. While recent autoregressive models exist with stronger results in English, mT5 remains one of the most comprehensive models supporting numerous languages. We fine-tune using a subset of \datasetname{} for the 101 languages supported by mT5, called \modelname. Our evaluation encompasses both multilingual Natural Language Understanding (NLU) and open-ended generation  (NLG).

\subsection{Baselines}
We compare our \modelname{} model against four
state-of-the-art multilingual instruction-following models.

 \textbf{mT0}
      \cite{muennighoff-etal-2023-crosslingual}:  An mT5-XXL-based model, instruction-tuned using the xP3 dataset,
      which consists of 16 reformulated NLP tasks, including
      summarization, QA, and classification for 46 languages.

    \textbf{Okapi} \citep{lai2023okapi}: A series of language-specific instruction-following models based on Bloom-7b \citep{workshop2023bloom} and Llama-7b \citep{touvron2023llama}. Each model is independently fine-tuned on translations of English synthetic data, followed by preference optimization for a specific language.

   \textbf{mT0x}: An mT5-XXL model instruction-tuned using
   the extended xP3 dataset, xP3x, covering 101 languages
   \citep{ustun2024aya}, providing a fair comparison
   regarding the number of languages.
   
   \textbf{Aya-101} \citep{ustun2024aya}: Uses xP3x, translated Aya Collection, a subset of DataProvenance \citep{longpre2023data} and translated ShareGPT-Command to instruction-tune mT5-XXL for 101 languages.

\subsection{Training Details}

Our experiments utilize the TPU Research Cloud (TRC) program, employing a TPU v4-32 with 32 chips and the T5X framework from Google. We set both input and output lengths to 1024 tokens and implement data packing. The effective batch size is 64, achieved through gradient accumulation (batch size of 8 with 8 accumulation steps). Following \citet{ustun2024aya}'s findings, we use a fixed learning rate of 3e-4 without a scheduler and trained for 5 epochs. For generation tasks, we apply nucleus sampling with top\_p=0.8 and temperature=0.9, as per \citet{Holtzman2020The}.

\begin{figure*}[t]
	\centering
	\includegraphics[width=\linewidth]{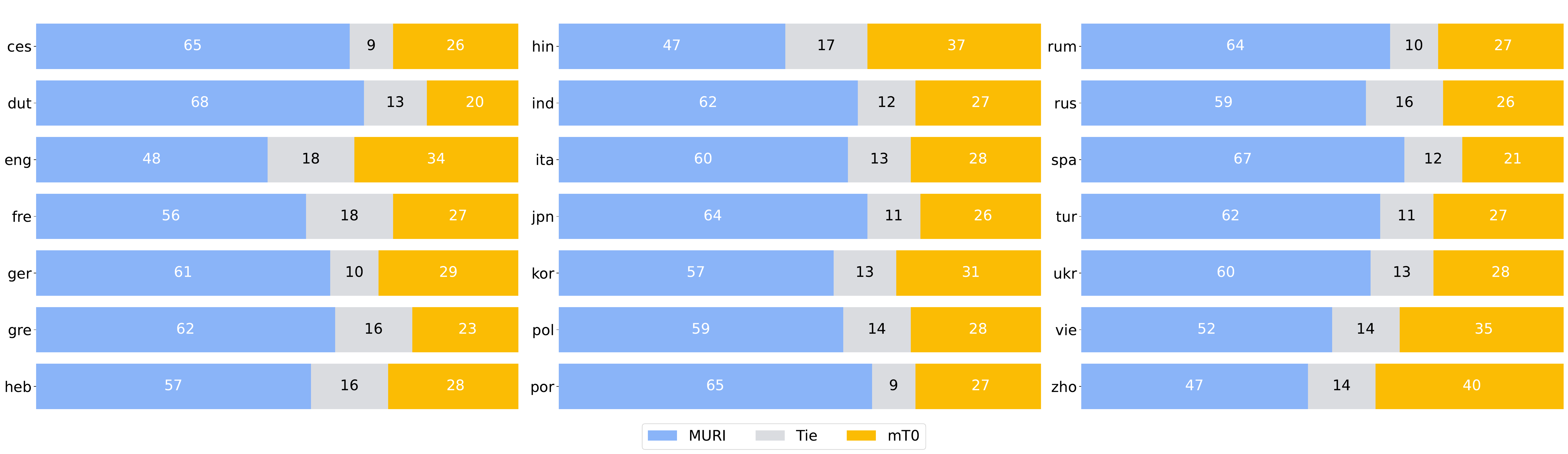}
	\caption{Win rates \modelname{} vs.\ mT0 outputs across
		21 languages on
		TranslatedDolly,
		the translated Dolly dataset. Win rates are determined by Command R+  as judge. The average win rates are 59\% and 28\% for \modelname and mT0, respectively.}
	\label{fig:winrates}
\end{figure*}

\subsection{Evaluation Benchmarks}
We evaluate the models in both multilingual and monolingual
settings for NLU and open-ended generation tasks. Two
evaluations use \textbf{TranslatedDolly} \citep{singh2024aya}, a
translated version of Dolly \citep{conover2023free}, a
human-annotated English instruction-tuning dataset.

    \textbf{Multilingual settings.}
         \emph{NLU}: Multilingual MMLU \citep{lai2023okapi}
         dataset, created by translating the English MMLU
         dataset to \mmmlunumlangs{} languages. We evaluate
         using the lm-evaluation-harness framework \cite{lmevaluationharness}
         with a 5-shot setup.\\
        \emph{NLG}:
TranslatedDolly,
evaluated on 21 languages using the multilingual Command R+
\cite{commandrplus}
model as an LLM judge.

 \textbf{Monolingual low-resource settings.}
 \emph{NLU}: Taxi1500 \citep{ma2023taxi1500} for classification with a 6-shot setup based on a parallel Bible corpus covering 1500 languages.\\
        \emph{NLG}:
TranslatedDolly

\section{Multilingual Model Evaluation}

We first evaluate our model \modelname{} on the few-shot
multilingual MMLU task. Table \ref{tab:m_mmlu} shows 
that \modelname{} clearly outperforms previous 
models (Okapi, mT0, mT0x), with an average relative improvement
of more than \textbf{14.3\%} (from 31.5 to 36.0). 
\modelname{} consistently outperforms prior 
models across all languages, with the exception of Aya-101.

While Aya-101 shows slightly better performance than
\modelname{} in NLU, we note that  Aya-101
is the result of a computationally-heavy training
process involving around 25 million samples. This includes a
lot of
translated data (47.5\% of the
training mixture) and
data
synthetically generated and translated
based on the ShareGPT dataset using a proprietary model
(22.5\% of the
training data). Thus, around 60\% of their training
data relies on translation which may introduce systematic 
translation artifacts known as translationese 
\citep{gellerstam1986translationese,yu2022translate} 
in the model outputs. 
However, this effect is difficult to evaluate with 
current metrics.  Given these factors, we
primarily compare \modelname{} with mT0 in NLG.

For NLG evaluation, we compare \modelname{} with mT0 on
TranslatedDolly
\citep{singh2024aya} and compare outputs using the
multilingual Command R+ model as a judge. From
TranslatedDolly, we select
the 21 languages that Command R+ supports.
Figure \ref{fig:winrates} shows that \modelname{}
consistently outperforms mT0 across all languages. Also across
all languages, \modelname{}'s win rate against mT0 is 59\%,
with lose and tie rates of 28\% and 13\%.

The lowest improvement in NLG is
for simplified Chinese, with a 47\% win rate vs.\
40\% loss rate. We hypothesize that code-switching within
different varieties of Chinese (as discussed in
\S\ref{sec:quality_assessment}) contributes to
this limited improvement.

\begin{table*}[ht]
  \centering
  \footnotesize
\begin{tabular}{lccccccccccc}
\toprule
\textbf{Language} & \textbf{aze} & \textbf{bel} & \textbf{bul} & \textbf{cym} & \textbf{gla} & \textbf{kaz} & \textbf{khm} & \textbf{lao} & \textbf{slk} & \textbf{slv} & \textbf{Avg.} \\
\midrule
mT5 & 20.4 & 22.4 & 20.7 & 18.4 & 19.3 & 19.8 & 16.5 & 21.3 & 19.2 & 18.9 & 19.7 \\
\textsc{Aya}\textsubscript{1} & 37.0 & 32.1 & 34.4 & 33.0 & 28.7 & 44.7 & 30.0 & 32.7 & 38.1 & \textbf{40.3} & 35.1\\
\textsc{Aya}\textsubscript{1}+\monomodelname & \textbf{39.5} & \textbf{33.7} & \textbf{38.1} & \textbf{35.5} & \textbf{35.2} & \textbf{46.7} & \textbf{31.3} & \textbf{33.0} & \textbf{39.1} & 39.6 & \textbf{37.2} \\
\bottomrule
\end{tabular}
\caption{Monolingual NLU performance on the Taxi1500 classification task across different low-resource languages. Scores are accuracy using a 6-shot setup.}
\label{tab:performance_comparison}
\end{table*}

\begin{table}[t!]
    \centering
    \resizebox{0.9\linewidth}{!}{
    \begin{tabular}{lcclcc}
    \toprule
    & \textsc{Aya}\textsubscript{1} & \textsc{Aya}\textsubscript{1}
 &   & \textsc{Aya}\textsubscript{1} & \textsc{Aya}\textsubscript{1} \\
    &  & +\monomodelname
  &  &  & +\monomodelname \\
    \midrule
    \textbf{aze} & 4\% & 4\% &    \textbf{kaz} & 2\% & 3\% \\
    \textbf{bel} & 3\% & 6\% &    \textbf{khm} & 2\% & 4\% \\
    \textbf{bul} & 6\% & 7\% &    \textbf{lao} & 3\% & 1\% \\
    \textbf{cym} & 2\% & 4\% &    \textbf{slk} & 4\% & 2\% \\
    \textbf{gla} & 2\% & 4\% &    \textbf{slv} & 6\% & 3\% \\
    \bottomrule
    \end{tabular}
    }
    \caption{Win rate comparison of  Aya\textsubscript{1} and Aya\textsubscript{1}+\monomodelname models vs.\ gold human outputs across different low-resource languages in NLG. The average win rates are 3.4\% and 3.8\% for \textsc{Aya}\textsubscript{1} and \textsc{Aya}\textsubscript{1} + \monomodelname, respectively.}
    \label{tab:winrate_comparison_mono}
\end{table}
\enote{AI}{should we add an average row?}

\section{Monolingual Evaluation in Low-Resource Setting}
To evaluate the capabilities of \datasetname and  Aya
in low-resource settings, we
conduct an additional set of
experiments with only monolingual training. We first
select ten low-resource languages: Azerbaijani,
Kazakh, Lao, Khmer, Welsh, Scottish Gaelic, Belarusian,
Bulgarian, Slovenian, and Slovak. While available in Aya,
these languages are not part of the human-annotated
portion of Aya and only have examples via translation, thus
possibly lacking in cultural context and
idiomaticity.
We test in our
experiment how well \datasetname{} complements translated
content in this setting. Furthermore, the languages were
chosen to represent diverse language families: Turkic,
Tai-Kadai, Austroasiatic, Celtic, and Slavic.

For this low-resource scenario, we sample at most 15K
examples from both Aya and \datasetname{}. Then we
instruction-tune mT5-XXL for each language and
for Aya and Aya+\datasetname{} separately, resulting in Aya\textsubscript{1} and Aya\textsubscript{1}+\monomodelname models.

Since many of these languages are not supported by
multilingual MMLU and Command R+, we use
the few-shot
classification task
Taxi1500
\citep{ma2023taxi1500}
for NLU. For NLG, we use
TranslatedDolly; however, we translate model
outputs to English (via Google Translate) and calculate win
rates with Llama-3-70B-Instruct
of translated outputs vs.\
Dolly's gold English
human outputs.

Table \ref{tab:performance_comparison} shows that
incorporating \datasetname{} consistently improves performance for
low-resource languages, except for Slovakian. The
baseline mT5 has 19.7\% accuracy (slightly
above random chance: 16.7\%) while Aya has
35.1\%. Even though Aya's performance is impressive, we
observe that incorporating \datasetname{} further improves
the results to 37.2\%. This shows that \datasetname{} can
complement Aya in low-compute and low-resource settings and
can further improve its performance.

Table
\ref{tab:winrate_comparison_mono}
shows that on average, the win rate
of Aya\textsubscript{1}
is 3.4\% and
of Aya\textsubscript{1}+\monomodelname
is 3.8\%.
This indicates that the
models  do not produce good-quality
outputs for these low-resource languages. While both Aya-101 and
\modelname{} demonstrate better NLG performance than prior
multilingual instruction-tuning models such as mT0, this
shows that current models are still limited in their NLG
capabilities for low-resource languages.

We hypothesize that
the limitations of our base model,
mT5, make it hard to achieve large improvements in NLG for
low-resource languages.
As recent autoregressive models begin
to support a larger number of languages, we anticipate that
\datasetname{}, with its human-written outputs, will be used
effectively to improve NLG performance for
low-resource languages.

\section{Conclusion}

This study presents \approachname (\approachnameshort), a novel approach for generating high-quality instruction tuning datasets for low-resource languages. Our method addresses limitations of translation-focused multilingual datasets by using human-written texts as outputs, combined with a translation pipeline and LLMs to create contextually appropriate instructions. The resulting dataset, \datasetname, of more than 2 million pairs across 200 languages greatly expands the resources available for multilingual language models.

Evaluation by native speakers from 13 languages confirmed the dataset's quality and idiomaticity. Our instruction-tuned mT5-XXL model, \modelname{}, strongly outperformed previous models on NLU and NLG in both multi- and monolingually. Notably, incorporating \datasetname{} improved performance for most low-resource languages, effectively complementing existing datasets like Aya.

While challenges remain, particularly in NLG for
low-resource languages, \datasetname{} represents a an important step towards more inclusive and linguistically diverse language models. Future work will focus on refining data quality and leveraging advanced multilingual models to further improve performance across languages.

\section{Limitations}
\raggedbottom
Despite the promising results obtained, several limitations
must be acknowledged in this study. First, we did not
perform clustering -- in contrast to \citeauthor{koksal2024longform}
-- due to uncertainties regarding the performance of multilingual encoders. Clustering could potentially enhance content diversity, ensuring a greater variety of linguistic and cultural contexts.

Additionally, the quality of the data can be further improved through more rigorous cleaning such as the removal of headers and footers from documents. Similarly, the \approachname{} methodology, particularly for low-resource languages, would benefit from more standardized source data. Our evaluation, involving native speakers, noted deficits in languages with less standardized orthography or prominent regional dialects. Additional preprocessing could address this issue.

Addressing these limitations in future work will involve integrating advanced clustering algorithms, enhancing data cleaning protocols, and expanding the dataset to include a wider range of languages.

\newpage
\bibliography{tacl2021, anthology_1, anthology_2}
\bibliographystyle{acl_natbib}

\section{Appendices}

\subsection*{Few-shot example used for Reverse Instruction generation}

\begin{table}[H]
	\centering
	
	\resizebox{\linewidth}{!}{
		\begin{tabular}{p{\linewidth}}
			\toprule
			
			\textbf{Answer:} Apache Kafka is a distributed system.
			The main components of Apache Kafka [...]\\
			\textit{> What kind of instruction could this be the answer to?} \\
			\textbf{Instruction:} What are the main components of Apache Kafka? \\
			\midrule
			\emph{(three more few shot examples)}\\
			\midrule
			
			\textbf{Answer:} [DOC]
			
			\textit{> What kind of instruction could this be the answer to?} \\
			\textbf{Instruction:} \\
			\bottomrule
		\end{tabular}
	}
	\caption{Few-shot examples used for reverse instruction generation.}
	\label{tab:few_shot_prompt}
\end{table}

\subsection*{Guideline for Evaluating \datasetname}

\begin{tcolorbox}[colback=blue!5!white,colframe=blue!75!black, title=Task Description]
You will be presented with a series of 30 instructions (prompts) and corresponding outputs (answers) based on them. Your task is to evaluate the instruction-output pairs based on several attributes to determine their quality and effectiveness in guiding a Large Language Model toward generating appropriate outputs.
\end{tcolorbox}

\begin{tcolorbox}[colback=yellow!5!white,colframe=yellow!75!black, title=Example Instruction-Output Pair]
	\textbf{Instruction:} What is a fracture? \\
	\textbf{Output:} A fracture is the (local) separation of a body into two or more pieces under the action of stress.
\end{tcolorbox}

\subsection*{Attributes for Evaluation}

\begin{enumerate}[leftmargin=0.5cm,labelsep=0.5cm]
	\item \textbf{Alignment:} Determine whether the instruction aligns with the output on a scale of 1 to 5, where:
	\begin{itemize}
		\item \textbf{1:} The instruction and the output are completely misaligned, making it difficult to understand how the output was generated based on the given instruction (e.g., the output does not or not fully answer the instruction).
		\item \textbf{5:} The instruction and the output are perfectly aligned, providing clear guidance on how to generate the output based on the given instruction.
	\end{itemize}
	
	\item \textbf{Instruction Format:} Identify if the instruction is phrased as an instruction or question:
	\begin{itemize}
		\item Mark as "Instruction" if the given instruction provides a directive for generating the response, e.g., it is phrased as an instruction or question.
		\item Mark as "No Instruction" if the given instruction is phrased as a statement, prompting no further answer.
	\end{itemize}
	
	\item \textbf{Grammatical and Lexical Correctness and Cohesiveness of the Instruction:} Assess whether the instruction is grammatically and lexically correct on a scale of 1 to 5, where:
	\begin{itemize}
		\item \textbf{1:} The instruction contains numerous grammatical errors and uses inappropriate or unclear language, hindering comprehension and interpretation. The text is not cohesive; parts of the text don’t belong together.
		\item \textbf{5:} The instruction is grammatically flawless and employs precise and appropriate language, facilitating clear understanding and interpretation.
	\end{itemize}
	
	\item \textbf{Grammatical and Lexical Correctness and Cohesiveness of the Output:} Assess whether the output is grammatically and lexically correct on a scale of 1 to 5, where:
	\begin{itemize}
		\item \textbf{1:} The output contains numerous grammatical errors and uses inappropriate or unclear language, hindering comprehension and interpretation. The text is not cohesive; parts of the text don’t belong together.
		\item \textbf{5:} The output is grammatically flawless and employs precise and appropriate language, facilitating clear understanding and interpretation.
	\end{itemize}
	
	\item \textbf{Informational Sufficiency:} Assess whether each instruction provides sufficient information for generating comprehensive outputs and whether it can be reasonably answered based on the provided information on a scale of 1 to 5, where:
	\begin{itemize}
		\item \textbf{1:} The instruction lacks essential information and details, making it impossible to generate a reasonable answer or is ambiguous and not understandable. \textbf{Example:} Summarize the article.
		\item \textbf{5:} The instruction provides ample information, and it is possible to be answered by a Large Language Model. \textbf{Example:} What does the word Rigadon mean?
	\end{itemize}
\end{enumerate}

\begin{tcolorbox}[colback=green!5!white,colframe=green!75!black, title=Annotation Instructions]
	\begin{enumerate}[leftmargin=0.5cm,labelsep=0.5cm]
		\item Read both the instruction and its output carefully.
		\item Evaluate each instruction and its output independently based on the provided attributes.
		\item Provide honest and thoughtful assessments for each instruction.
		\item The output can contain additional information typical of websites such as links or footers; please account for this in your assessment by penalizing Grammaticality or Alignment accordingly.
		\item If you encounter any difficulties or uncertainties, please refer back to these guidelines or reach out for clarification.
		\item You can add your own notes in the Notes column if you want to explain your evaluation.
	\end{enumerate}
\end{tcolorbox}

\end{document}